# Modular Neural Computer

**Florin Leon**

Department of Computers
Faculty of Automatic Control and Computer Engineering
"Gheorghe Asachi" Technical University of Iași, Romania
Email: florin.leon@academic.tuiasi.ro

**Abstract**

This paper introduces the Modular Neural Computer (MNC), a memory-augmented neural architecture for exact algorithmic computation on variable-length inputs. The model combines an external associative memory of scalar cells, explicit read and write heads, a controller multi-layer perceptron (MLP), and a homogeneous set of functional MLP modules. Rather than learning an algorithm end to end from data, it realizes a given algorithm through analytically specified neural components with fixed interfaces and exact behavior. The control flow is represented inside the neural computation through one-hot module gates, where inactive modules are inhibited. Computation unfolds as a sequence of memory transformations generated by a fixed graph. The architecture is illustrated through three case studies: computing the minimum of an array, sorting an array in place, and executing A* search on a fixed problem instance. These examples show that algorithmic procedures can be compiled into modular neural components with external memory while preserving deterministic behavior and explicit intermediate state.

## 1. Introduction

Neural architectures with external memory, most notably the Neural Turing Machine (NTM) [1] and the Differentiable Neural Computer (DNC) [2], offer an appealing model for algorithmic computation. They combine a neural controller with differentiable read and write operations over memory, which in principle allows them to represent iteration, variable binding, and intermediate state over inputs of varying length. Yet there remains a clear gap between this promise and the behavior observed in practice. On relatively simple algorithmic tasks, learned solutions often depend heavily on the training process, require large numbers of examples, and may fail to generalize reliably to inputs longer than those seen during training. In our own experiments, for example, a DNC can learn to compute the minimum of a short array, but performance does not extend robustly to substantially longer arrays.

In this paper, instead of learning an algorithm from data, we design a neural system that implements a given algorithm exactly. Another goal is computation on variable-length inputs, including long inputs, within a memory-bounded setting. The proposed architecture remains neural: it is built from feedforward multi-layer perceptron (MLP) modules, explicit read and write heads, and an external associative memory. However, the relations among these components are specified analytically rather than learned end to end.

Each MLP module implements an exact map corresponding to one step of an algorithm. A separate MLP controller emits the control signals and memory addresses required at each step. Computation then unfolds through repeated application of the same graph over memory states. The result is a modular neural program in which correctness follows from the design of the modules and their interfaces, rather than from statistical training alone.

This approach is less general than end-to-end differentiable memory systems, but it offers a different advantage: exactness. It shows that algorithmic procedures for variable-length inputs can be compiled into neural components with external memory while preserving precise, deterministic behavior.

The remainder of the paper is organized as follows. Section 2 reviews related work on memory-augmented neural networks and neuro-symbolic approaches to algorithmic computation. Section 3 presents the Modular Neural Computer (MNC) architecture, including its design principles, memory model, controller, module interfaces, and gating mechanism. Section 4 describes three case studies that illustrate the approach on increasingly richer algorithms: reduction, in-place transformation, and explicit search. The last section includes the conclusions.

The implementations used in the experiments are available at: https://github.com/florinleon/ModularNeuralComputer.

## 2. Related Work

Research on neural models with explicit memory was shaped by architectures that combined a neural controller with an addressable external memory. Neural Turing Machines [1] introduced differentiable read and write heads over memory and demonstrated learned behaviors such as copying, sorting, and associative recall. End-To-End Memory Networks [7] developed a related idea through recurrent attention over external memory, especially for multi-hop reasoning. Differentiable Neural Computers [2] extended this line with richer allocation and temporal-link mechanisms, while Sparse Access Memory [5] addressed the scalability limits of dense memory access.

A parallel line of work studied neural systems that learn algorithmic behavior or program execution from data. Pointer Networks [10] showed how attention can act as a position selector over variable-length inputs, which made them relevant to tasks such as sorting and combinatorial optimization. Neural Programmer-Interpreters [6] made program decomposition explicit through callable subprograms and supervised execution traces. Neural GPUs [4] demonstrated that weight-sharing architectures can learn arithmetic-style procedures and sometimes generalize to longer inputs than those seen in training. These approaches differ architecturally, but they share the goal of learning procedures from examples.

More recent work has shifted attention toward explicit neural execution of algorithmic steps. Neural Execution of Graph Algorithms [9] trained graph neural networks to imitate intermediate states of classical graph algorithms rather than only their final outputs. The CLRS Algorithmic Reasoning Benchmark [8] provided a broad common testbed spanning sorting, searching, dynamic programming, graph algorithms, string algorithms, and geometry. A Generalist Neural Algorithmic Learner [3] then examined whether a single neural processor can execute many different algorithms within a shared representation space.

The present work is closest in spirit to this literature, but departs from it in a fundamental methodological respect. The models above are principally designed to learn algorithmic computation from data and are typically evaluated by extrapolation beyond the training regime. The Modular Neural Computer is designed for exact algorithm execution. Its modules, interfaces, and control relations are specified analytically, and the external memory remains explicit and inspectable. In that sense, it is less general than learned memory-augmented or algorithmic-reasoning systems, but it offers stronger guarantees and a more direct correspondence between symbolic program phases and neural computation.

## 3. Modular Neural Computer Architecture

### 3.1. Design Philosophy

The architecture is built around the premise that an algorithm can be decomposed into a small number of exact transformations, and these transformations can be implemented as neural modules with a common interface. The goal is not to train a network to discover a procedure, but to realize a known procedure in a neural form that remains explicit, modular, and exact.

This leads to four design choices. First, the memory is external. Intermediate state is not hidden inside recurrent activations, but stored in an associative memory that can be read and updated through explicit operations. This allows the same mechanism to operate over inputs of varying length.

Second, the functional part of the system is modular. Rather than encoding all phases of a computation in a single network, the architecture uses a set of modules, each responsible for one algorithmic operation. This separation mirrors the internal organization of ordinary programs, where initialization, update, and termination are distinct transformations even when they act on the same state.

Third, all functional modules are homogeneous at the interface level. Each receives the same number of read values and produces the same number of write values. The semantic differences among modules are therefore expressed only by their internal maps. If a module does not require one of its inputs, that input is neutralized by zero weights. This preserves a clean system description and avoids specialized exceptions.

Fourth, control flow is represented inside the neural computation itself. The architecture does not rely on symbolic dispatch, programmatic branching, or external selection of the active module. Instead, the controller emits gate values, and these gates determine which module is active, i.e., contributes a nonzero output, at a given step. The result is a fixed computational graph whose behavior changes only through values carried in memory and in the controller output.

**3.2. Components and Interfaces**

*External Associative Memory*

The system uses an external associative memory whose cells contain scalar values. Let the address space be a fixed finite set $A = 0, \ldots, S-1$. At step *t*, the memory state is a scalar map $M_t : A \to \mathbb{R}$, or equivalently a value vector $V_t \in \mathbb{R}^S$, where $V_t[a] = M_t(a)$. Some addresses are reserved for control state, such as the current index, the input length, constants, or halting flags, while others store task data and intermediate results. The scalar choice is deliberate. It keeps the memory interface minimal, so each read head returns one scalar and each write head updates one scalar cell. This makes the computational state explicit and stepwise program invariants easy to state.

The memory is associative in the sense that access is mediated by keys rather than by direct array indexing inside the execution logic. In the present implementation, each address $a \in A$ has a canonical key $e_a \in \mathbb{R}^S$, the corresponding one-hot basis vector. The key matrix is therefore $K = I_S$, so the key space and the address space coincide. This choice is intentionally simple: there is no learned content-based key dictionary and no compression of addresses into latent vectors. The addressing mechanism is explicit and transparent. The controller emits scalar addresses, these addresses are converted into key vectors, and the same key-based rule is used for both reading and writing. The value dimension is one in the current program, so the implemented memory is an associative array of scalar cells rather than a set of higher-dimensional records.

More precisely, if the controller produces a scalar address $q_t$, the memory first maps it into the shared key space through a function $\phi$. For an integer address $q_t = a$, one has $\phi(q_t) = e_a$. For a non-integer address $q_t \in [\ell, \ell+1]$, the implementation uses linear interpolation between adjacent keys:

$$\phi(q_t) = (\ell + 1 - q_t)e_\ell + (q_t - \ell)e_{\ell+1}. \tag{1}$$

Thus legal discrete addresses correspond to exact basis vectors, while fractional addresses produce convex combinations of multiple keys.

A read from address $q_t$ is then performed by comparing $\phi(q_t)$ with all memory keys and normalizing the resulting scores through a temperature-scaled softmax:

$$\omega_t = \text{softmax}\left(\frac{K\phi(q_t)}{\tau}\right), \tag{2}$$

$$\text{Read}(q_t) = \omega_t^\top V_t. \tag{3}$$

Because $K = I_s$, the score vector is simply the key vector itself. In the current implementations the temperature is $\tau = 10^{-4}$, so when $q_t$ is an integer address, the distribution $\omega_t$ is very concentrated on a single slot. The read therefore behaves almost exactly like a hard lookup. When $q_t$ is fractional, the read interpolates smoothly between multiple cells. This is why the implementation is best understood as an almost-discrete associative memory: it preserves the semantics of ordinary slot access on legal states, while retaining a continuous neural interface.

Writes use the same addressing rule. If $q_t$ is the selected write address and $v_t \in \mathbb{R}$ is the scalar to be stored, the implementation forms the same attention vector $\omega_t$ and updates each cell according to:

$$M_{t+1}(a) = \alpha \cdot \omega_t(a) \cdot v_t + (1 - \alpha \cdot \omega_t(a)) \cdot M_t(a), \tag{4}$$

with $\alpha = 1$ in the current implementations. For an integer address and the very small temperature used here, this is effectively an overwrite of a single slot, with all other cells left unchanged up to negligible numerical spillover from the softmax. For fractional addresses, the write is distributed across multiple cells in the same way that reads are interpolated across them. The implementation also includes a compatible soft delete operation, which attenuates the highly attended cell toward 0.

This memory design fits the execution model of the architecture. The controller emits a number of $n_r$ read addresses (here $n_r = 3$) and a number of $n_w$ write addresses (here $n_w = 2$) from a fixed control input. The read heads retrieve $n_r$ scalars, and in fact all modules receive those same scalars. The active module produces $n_w$ output values, and those values are written back to the controller-selected locations. The memory is therefore the explicit carrier of algorithmic state. In the minimum case study presented in section 4.1, for example, it stores both immutable input values and mutable program variables such as the current index and the running minimum. Computation then appears as a sequence of exact transformations of memory cell values, not as an evolution of hidden activations.

*Controller*

The controller is an MLP that receives a fixed control input extracted from memory. In the present design, the control input consists of $n_r$ scalar values $c_t = (c_t^1, ..., c_t^{n_r})$, typically including the current execution index, the problem size, and sometimes the address of a constant 0 cell. The controller does not perform memory reads internally. Its input is supplied through a dedicated control-read interface, which remains the same at every step.

From this input, the controller produces two categories of outputs. The first is a gate vector $g_t = (g_t^1, ..., g_t^K)$, where $K$ is the number of functional modules. This vector is one-hot: exactly one component is equal to 1 and all others are 0. The second category consists of read and write addresses $r_t = (r_t^1, ..., r_t^{n_r})$ and $w_t = (w_t^1, ..., w_t^{n_w})$. These addresses specify which memory cells will provide the functional inputs and which cells will receive the outputs at the current step.

The controller therefore serves two roles: it identifies the current algorithmic phase, and it determines the locations in memory that are relevant for that phase. Both roles are realized by the same MLP, with no auxiliary symbolic logic.

*Memory Interface*

Given the controller outputs $r_t$, the system performs $n_r$ functional reads from memory: $x_t = (x_t^1, ..., x_t^{n_r}) = (M_t(r_t^1), ..., M_t(r_t^{n_r}))$. These $n_r$ scalars are broadcast to all modules. Every module therefore receives the same read values at step $t$. The distinction among modules lies not in the data they receive, but in how they transform those data when their gate is active.

*The Set of Functional Modules*

The functional core consists of a set of $K$ MLP modules $\mathcal{F}^1, ..., \mathcal{F}^K$ with the same interface: $\mathcal{F}^k : (g_t^k, x_t^1, ..., x_t^{n_r}) \mapsto (u_t^{k,1}, ..., u_t^{k,n_w})$. Thus each module receives one scalar gate and $n_r$ scalar data inputs, and produces $n_w$ scalar outputs. This interface is intentionally uniform. A module may depend on all read values, on only a subset of them, or on one read value together with the gate. Such differences are internal to the module. The surrounding architecture remains unchanged.

*Producing Output Values*

After all modules have produced their outputs, the final write values are obtained by componentwise addition:

$$y_t^i = \sum_{k=1}^{K} u_t^{k,i}. \tag{5}$$

These values are then written to the controller-selected addresses:

$$M_{t+1}(w_t^i) \leftarrow y_t^i. \tag{6}$$

All other memory cells remain unchanged unless the particular task definition specifies otherwise.

*Gating and Module Inhibition*

The gate values are not soft attention weights and are not used to scale candidate outputs after a computation step. They are part of the input to each module, and each module is constructed so that when the gating signal $g_t^k = 0$, its outputs are automatically 0. When $g_t^k = 1$, the module applies its task-specific transformation to the relevant read values.

This point is central to the execution model. All modules are present at every step, and they all receive the same read inputs. However, only the gated module produces a meaningful output. The remaining modules are inhibited internally and return 0. Because the controller gate is one-hot on valid states, the additive merge of module outputs does not mix competing transformations. It simply passes through the unique nonzero result.

This design avoids two alternatives that would be less satisfactory. The first is symbolic dispatch, where an external program selects one module and skips the others. The second is post hoc cancellation, where multiple modules produce outputs and a later stage suppresses the irrelevant ones. In the present architecture, inhibition is local to each module and is part of the module definition itself.

*Component Interaction During Execution*

For an execution step, the system first retrieves the control values required by the controller. The controller then emits a one-hot gate over the set of modules and the read and write addresses for the current step. The read heads use these addresses to fetch the scalar values from memory. These values, together with the module-specific gate signals, are fed to all modules. Exactly one module produces a nonzero pair of outputs, while the others return 0. The outputs are added, written back to memory at the selected addresses, and the updated memory becomes the state for the next step.

The same computation graph is reused at every step. What changes from one step to the next is the memory state and, through it, the controller output. In this way, algorithmic progression is represented as a sequence of exact memory transformations generated by a fixed modular neural system.

## 4. Case Studies

### 4.1. Finding the Minimum Value of an Array

We now illustrate the architecture for the problem of computing the minimum of a finite array of scalars. Given an input array $a_1, \ldots, a_n$, the objective is to produce $\min(a_1, \ldots, a_n)$ exactly, for any valid length $n$ supported by the memory. This example is useful to pinpoint the essential components of the method while remaining simple enough to describe completely.

*Problem Formulation and the Symbolic Procedure*

The underlying symbolic procedure is the standard linear scan. The algorithm first copies the first array element into a dedicated memory cell that stores the current minimum. It then visits the remaining elements one by one, replacing the stored value whenever a smaller element is encountered. After the final element has been processed, the cell holding the running minimum already contains the exact result.

In symbolic form, the procedure can be written as:

- Initialize $m \leftarrow a_1$;
- for $i = 2, \ldots, n$, set $m \leftarrow \min(m, a_i)$;
- return $m$.

The neural implementation preserves this decomposition. It uses 3 functional modules corresponding to initialization, iterative update, and termination. The purpose of the controller in this case is to determine which of these 3 phases is active at a given step and to expose the appropriate memory values to the modules.

*Memory Layout*

The minimum task requires only a small number of reserved memory locations in addition to the cells that store the input array. One contiguous region of memory contains the array values $a_1, \ldots, a_n$. Separate reserved addresses store the current index $i$, the array length $n$, the current minimum $m$, a constant 0 value, and a running flag or equivalent halting indicator. The exact numerical encoding of addresses is not important at the conceptual level; what matters is that these addresses are fixed and known to the controller.

This layout separates data from control. The array cells are immutable input locations. The current index is mutable control state. The current minimum is mutable task-specific state. As a result, each step of execution can be understood as a transformation of a small number of scalar memory cells.

*Controller Specialization*

For this task, the controller receives as control input the triple ($i$, $n$, 0), where $i$ is the current array index, $n$ is the array length, and 0 is the value stored in the constant zero cell. From this input it emits 3 gate values ($g_{init}, g_{update}, g_{stop}$), together with 3 functional read addresses and 2 write addresses.

The gate values encode the current phase. On legal states, the controller satisfies:

- $g_{init} = 1$ if and only if $i = 1$;
- $g_{update} = 1$ if and only if $2 \leq i \leq n$;
- $g_{stop} = 1$ if and only if $i = n + 1$.

All other gate values are 0. Thus the controller converts the scalar control state into an exact one-hot phase indicator.

The address outputs are phase dependent. During initialization, the controller reads the first array element and the current index, and writes the resulting minimum and the incremented index. During the update phase, it reads the current minimum, the current array element $a_i$, and the current index, and writes back the updated minimum and the incremented index. During termination, it reads the current minimum and routes it to the designated output location or final state cell. The important point is that these addresses are produced directly by the controller MLP for the current state. No external rule translates phase labels into addresses.

*Functional Modules*

The minimum program uses 3 modules, all with the same interface. Each receives its own gate and the 3 scalars returned by the read heads. Each produces 2 scalar outputs. The semantic difference lies entirely in the map computed when the gate is active.

*Initialization Module*

The initialization module performs the first assignment of the running minimum. When its gate is active, it implements:

$$(x_1, x_2, x_3) \mapsto (x_1, x_2 + 1). \tag{7}$$

For this task, the controller supplies $x_1 = a_1$, $x_2 = i$, and $x_3 = 0$ during the initialization step. The module therefore writes the first array element into the minimum cell and increments the index from 1 to 2. The third read input is unused in this phase and is cancelled internally by zero weights. Its inhibited behavior is equally important: when the gate is 0, the module returns (0, 0) directly.

*Update Module*

The update module performs the core comparison step. When active, it implements:

$$(x_1, x_2, x_3) \mapsto (\min(x_1, x_2), x_3 + 1). \tag{8}$$

The controller arranges the reads so that $x_1$ is the current minimum, $x_2$ is the current array element $a_i$, and $x_3$ is the current index $i$. The module therefore computes the new running minimum and increments the index to the next array position.

This is the only module that uses all 3 read inputs. The exact minimum is realized through the identity:

$$\min(x_1, x_2) = \frac{x_1 + x_2 - |x_1 - x_2|}{2} \tag{9}$$

with an equivalent ReLU-based construction inside the MLP. No approximation is involved in this comparison.

As before, if the gate is 0, the module returns (0, 0).

*Stop Module*

The stop module handles the algorithm termination. It outputs a halting marker together with the final minimum. A representative map is:

$$(x_1, x_2, x_3) \mapsto (-1, x_1), \tag{10}$$

where $x_1$ is the current minimum and the remaining inputs are ignored. The first output may be written to a running-flag cell to indicate completion, while the second preserves or exposes the final result. Since the running minimum has already been computed during the update phase, the stop module performs no further comparison.

Again, when inhibited with a zero gate, the module returns (0, 0).

### *Execution*

The computation begins with $i = 1$. The controller activates the initialization module, which copies $a_1$ into the minimum cell and advances the index. For every subsequent step with $2 \leq i \leq n$, the controller activates the update module. Each such step compares the current minimum against the next array element and stores the smaller value back into memory. When the index reaches $n + 1$, the controller activates the stop module, which writes the final completion state. The result follows immediately from the memory invariant maintained by the update phase: after processing index $i$, the dedicated minimum cell contains $\min(a_1, \ldots, a_i)$.

This invariant is established at initialization for $i = 1$, preserved by every update step, and therefore yields $\min(a_1, \ldots, a_n)$ at termination.

This example demonstrates the intended use of the architecture in its simplest form. The controller realizes a phase schedule designed analytically. The modules do not approximate the relevant transformations; they implement them exactly. The memory holds explicit algorithmic state, including the current index and the running minimum. The resulting system is therefore a neural realization of a concrete algorithm, rather than a learned approximation.

**4.2. Sorting an Array**

The second case study extends the same neural program template to a task in which the data themselves are modified in place. Unlike the minimum problem, where a single running statistic is updated while the input array remains unchanged, sorting requires repeated local rewrites of the array cells. For this reason, the sorting instance differs from the minimum case not in its overall organization, which remains the same, but in the state variables, the controller logic, the module set, and the write interface of the active modules.

*Symbolic Procedure*

The underlying symbolic procedure is a pass-based adjacent sorting routine. At any moment, the computation is determined by two control variables: the current pair index $i$ and the current pass limit $p$. If $i < p$, the algorithm reads the adjacent pair $(a_i, a_{i+1})$, rewrites it in nondecreasing order, and advances to $i + 1$. If $i = p$ and $p > 1$, the current pass is complete: the pass limit is decremented and the pair index is reset to 1. If $i = p = 1$, the algorithm halts. This modularization separates pair processing, pass transition, and stopping into distinct routines.

This decomposition has an immediate neural consequence. In the minimum case, the main loop updates a single running value and one index. Here, one step may need to rewrite two array cells and one control cell simultaneously. The sorting realization therefore uses 3 write values per functional step rather than 2 to reflect the fact that an exact adjacent compare-and-swap step must update both array positions and also advance the traversal state.

*Memory State and Task-Specific Variables*

The array occupies the same kind of contiguous memory region as in the minimum case. The difference lies in the working state. Sorting requires one cell for the current pair index $i$, one cell for the current pass limit $p$, one permanent 0 cell, and one running flag. The array length $n$ is still stored explicitly, although once the initial pass is set to $p = n$, the active control state is carried by $(i, p)$ rather than by $(i, n)$. There is no dedicated set of cells for the final result; the sorted output is the array itself after the sequence of local rewrites has terminated. The memory therefore acts not only as control state and workspace, but also as the evolving data object being transformed step by step.

*Controller Specialization*

The controller reads a fixed triple from memory and emits exact phase gates together with the functional read and write addresses. The controller input is ($i$, $p$, 0). From this input it produces 3 one-hot gates: $(g_{process}, g_{next}, g_{stop})$. Their semantics are:

- $g_{process} = 1$ when $i < p$;
- $g_{next} = 1$ when $i = p$ and $p > 1$;
- $g_{stop} = 1$ when $i = p = 1$.

This differs from the minimum controller in two ways. First, phase selection no longer depends on comparing the current index to a fixed problem size alone. It depends on the relation between two mutable control variables, $i$ and $p$. Second, the stopping condition is not "past the end of the array", but "the final pass has length 1". The controller must therefore encode both traversal within a pass and transition between passes.

The address outputs also differ substantially from those in the minimum case. During pair processing, the controller reads $a_i$, $a_{i+1}$, and the current index $i$, and writes back the reordered pair together with the incremented index. During pass transition, it reads the current pass limit $p$ and writes the decremented pass limit together with the reset pair index 1. During stopping, it writes a negative value to the running-flag cell and leaves the remaining writes neutral. Thus, unlike the minimum controller, which writes either a running statistic or a halting flag plus a result cell, the sorting controller must route 3 write heads in a phase-dependent way.

*Functional Modules*

The sorting implementation uses 3 functional modules.

The first module performs adjacent pair processing. When active, it receives the left value $a_i$, the right value $a_{i+1}$, and the current pair index $i$. Its output is: $(\min(a_i, a_{i+1}), \max(a_i, a_{i+1}), i+1)$. The first two outputs overwrite the original pair in sorted order, and the third advances the current pair index. This module is the central computational step of the sorting program. It uses the same exact comparison primitives as in the minimum case, but now both extrema are needed: the smaller value is written back to the left cell and the larger value to the right cell.

The second module performs the pass transition. When active, it receives the current pass limit $p$ and outputs ($p - 1$, 1, 0). The first value shortens the next pass by one position, the second resets the current pair index to the beginning of the array, and the third is a neutral write. This module is specific to pass-based sorting, where progress is measured not only by movement along the array but also by contraction of the unsorted suffix.

The third module is the stop module. Its purpose is to terminate execution by writing a negative value to the running-flag cell. The sorted result is already present in the array cells, so no dedicated result propagation step is needed. Termination simply marks completion of the in-place transformation.

*Execution*

Sorting requires a full triangular number of steps, because every pass consists of several pair-processing steps followed by a pass-transition step. For pass limits $p = n, n-1, \ldots, 2$, the number of steps is $\sum_{p=2}^{n} p + 1 = n(n+1)/2$.

The correctness argument follows the usual invariant for adjacent-exchange sorting. Within a pass of limit $p$, each pair-processing step preserves the multiset of array values while moving larger values to the right locally. After completion of that pass, the largest value among the first $p$ positions occupies position $p$. The next-pass module then reduces the active prefix to $p - 1$, and the same reasoning applies recursively. When the final stop state is reached, the entire array is sorted in nondecreasing order.

What this case study adds, relative to the minimum example, is the demonstration that the same neural program scheme can express not only a reduction over an input sequence, but also an exact in-place transformation of the sequence itself.

**4.3. The A* Algorithm**

This case study extends the same modular neural program to a search procedure. In contrast with the previous examples, where the main data object is a fixed array, A* maintains a dynamically growing set of search nodes. The case is therefore useful for showing that the architecture can represent not only iterative transformations over a static input, but also the controlled construction and traversal of an explicit search tree in external memory.

The implementation considered here is the basic form of A* on a fixed problem instance. The graph, edge costs, heuristic values, start state, and goal state are all written into memory before neural execution begins. This initialization step is programmatic rather than neural. The search procedure itself is then executed by the same overall architecture as in the previous case studies: a controller MLP reads a fixed control state, emits one-hot module gates and memory addresses, and a set of gated functional modules performs the corresponding memory updates. In this case study, however, exactness is achieved in a problem-specific way. Because the problem instance is fixed and finite, the controller and module maps are compiled from an explicit specification of the valid execution states and transitions for that instance. The resulting runtime system is purely neural, but its weights encode a finite search procedure rather than a domain-independent A* solver.

*Program Organization*

The program is organized around 3 memory regions. The first stores the problem instance. For each graph state, memory contains its identifier, heuristic value, number of outgoing actions, and up to two successor-cost pairs. The second region stores the dynamic control state of the algorithm, including the current phase, the next free search-node slot, the current node scan position, the best node found so far during open-list scanning, the currently selected node, the current action index during expansion, the number of open nodes, and the final solution pointer. The third region stores the generated search nodes themselves as fixed-width records.

Each search-node record occupies a separate block of scalar memory cells. The fields include the graph state, the parent search-node index, the generating action, the path cost $G$, the heuristic $H$, the evaluation value $F = G + H$, an open flag, and a validity flag. The memory is intentionally sparse: blocks are separated by unused cells so that records are easy to inspect and their addresses are easy to compute. This reflects the aim of making the algorithmic content transparent rather than memory-efficient.

This case study therefore uses external memory in two different roles. One part of memory stores the description of the search problem, while another part stores the evolving search process itself. The first remains unchanged after initialization. The second grows as new search nodes are generated and is updated throughout execution.

As in the previous case studies, the runtime controller does not perform side computations outside the neural graph. It receives a fixed control input from memory and emits exact phase gates together with the read and write addresses needed at the current step. The functional part of the computation is realized by a homogeneous set of modules. All modules share the same interface, use the same number of read heads and write heads, and suppress irrelevant inputs and outputs through zero weights.

*Functional Modules*

The A* program uses 6 modules, which realize the standard phases of the basic A* algorithm. Each module is a pure MLP at runtime. Its weights are chosen so that, on the finite set of execution states relevant to the instance, it implements the required transition exactly.

The first module, *InitRoot*, creates the root search node. It reads the start state and its heuristic value from the problem-description region and writes the first search-node record with $G = 0$, $H = h(s_0)$, $F = h(s_0)$, null parent and action fields, and the open flag set to 1. It also initializes the control cells, including the number of open nodes and the next free record index. This module is the analog of the initialization module in the minimum and sorting cases, but here the initialized object is a full search-node record rather than a single scalar accumulator or a local array position.

The second module, *StartOpenScan*, initializes a scan over all generated search nodes. It sets the scan index to the first record, clears the current best-node pointer, and initializes the current best $F$ value to a large constant. This prepares the explicit search over the open list.

The third module, *ScanOpenNode*, processes one candidate record during that scan. Its role is to update the current best open node and then advance the scan index. Conceptually, it compares the candidate record against the current best record under the usual A* criterion: only valid open nodes are eligible, and among those the node with smallest $F$ is retained. This is the A* analog of the scan used in the minimum case, but applied to search-node records rather than raw array values. Because the set of relevant execution states is finite for the fixed instance, this comparison-and-update behavior can be compiled into an exact module map.

The fourth module, *FinishOpenScan*, completes the extraction of the next node to expand. It transfers the selected best node into the dedicated current-node control cells, clears its open flag, decrements the open-node count, and advances control to the next phase. In the event that no open node remains, the same module writes the failure state instead. As in the rest of the A* program, these alternatives are not implemented as runtime symbolic branches, but as outputs of the compiled module.

The fifth module, *GoalTest*, separates terminal from non-terminal selected nodes. On goal states, it writes the solution-node pointer and clears the running flag. On non-goal states, it initializes the action-expansion phase by setting the current action index and loading the number of outgoing actions for the selected graph state. Since the goal state is fixed as part of the problem description, this phase transition is again compiled into an exact module map for the problem instance.

The sixth module, *ExpandAction*, generates one child search node for the current action slot. It reads the selected node, the corresponding successor and edge cost from the problem-description region, and the heuristic value of the successor. It then writes a new search-node record with the successor state, parent pointer, generating action, updated path cost $G$, heuristic $H$, evaluation value $F$, and open flag set to 1. It also increments the next free record pointer, increments the open-node count, and advances the current action index. When the available actions for the current node have been exhausted, the module transitions control back to the open-list scan. This phase is the most visibly search-specific part of the program, since it realizes the growth of the search tree by appending new records to the dynamic node region of memory.

Together, these modules realize the standard unoptimized A* loop as a sequence of exact neural phases: initialize the root, find the open node with minimum $F$, test for goal, expand one outgoing action, and repeat.

The modular decomposition is slightly richer than in the previous case studies because A* contains 2 nested iterative processes: a scan over the open set to select the next node to expand, and a scan over the outgoing actions of that node to generate successors. The 6-module design isolates these two processes clearly. One subset of modules manages selection from the frontier, while another manages expansion of the selected node.

*The Actual Search Problem*

The graph used in this case study contains 7 states: *S*, *A*, *B*, *C*, *D*, *E*, *G*, where *S* is the start state and *G* is the goal state, as shown in Figure 1.

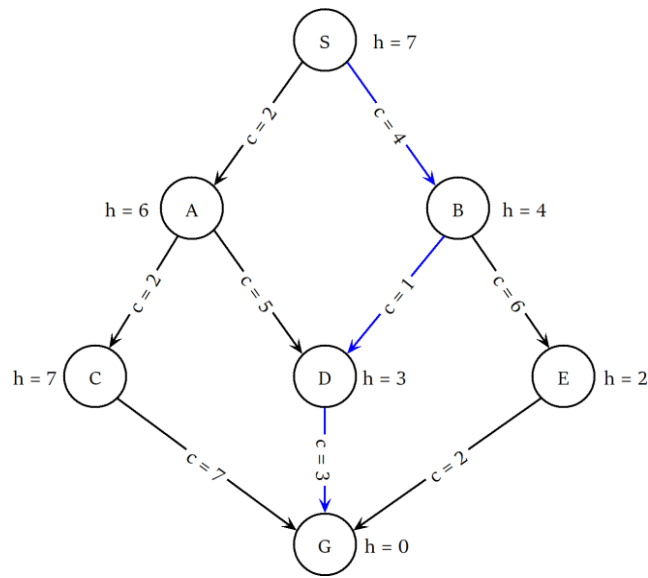

**Figure 1.** The problem solved by neural A* search

The costs of the directed edges are shown on the edges (e.g., $c = 2$), and the heuristic values are situated near the states (e.g., $h = 6$). This graph is small enough to inspect manually but rich enough to produce a nontrivial search tree. In particular, state *D* can be reached through 2 different partial paths, which means that the generated search tree contains distinct search nodes that correspond to the same graph state. Since this case study deliberately uses the unoptimized A* formulation without duplicate suppression or a closed list, that behavior is expected and provides a natural test of the explicit search-node representation.

*Neural Realization of the Solution*

The solution unfolds exactly as in ordinary A*. The root node for *S* is created first. The open-list scan then selects the currently open node with minimal *F*. That node is tested against the goal. If it is not a goal node, its outgoing actions are expanded one at a time into fresh search-node records. These new records enter the open list through their open flags, and the algorithm repeats. For the fixed graph above, the optimal solution path is $S \rightarrow B \rightarrow D \rightarrow G$ with a total cost of $4 + 1 + 3 = 8$. The final solution is therefore represented as an explicit chain of parent links through the generated search-node records.

This case study illustrates the main extension beyond the earlier examples. The minimum and sorting tasks operate on a fixed data object that is either scanned or rewritten in place. A* adds a second layer of dynamics: the algorithm must create new search-node records, maintain an explicit frontier, select the next record to expand, and preserve parent links for later path reconstruction. The same modular neural architecture can realize this process.

## 5. Conclusions

This work has presented the Modular Neural Computer (MNC) as a neural architecture for exact algorithmic execution with external memory. The central contribution is not a new training method, but a different use of neural components: MLPs, read-write heads, and associative memory are combined into a modular execution system whose correctness follows from construction. In this setting, the external memory is the explicit carrier of algorithmic state, and the neural modules act as local transformations over that state.

From a broader neuro-symbolic perspective, the proposed approach occupies an interesting middle ground. It preserves recognizably neural ingredients and a uniform computational interface, yet it does not rely on end-to-end statistical learning to induce the underlying procedure. In that sense, it differs both from classical symbolic program execution and from memory-augmented neural networks that emphasize learnability. The present approach instead emphasizes exactness, inspectability, and modular correspondence between program phases and neural components. These properties may be valuable in settings where reliability is important.

At the same time, this work also suggests a broader research direction. If exact modular neural programs can be constructed analytically for specific tasks, one may ask whether their modules and interconnections could be discovered automatically without reverting to standard end-to-end backpropagation. A promising possibility is an evolutionary or other population-based search over module definitions, interfaces, and controller-module couplings. Such techniques may offer a route toward neural program synthesis that remains modular and explicit, while avoiding some of the difficulties of pure gradient-based methods.